# A Market-Oriented Programming Environment and its Application to Distributed Multicommodity Flow Problems

**Michael P. Wellman**                    WELLMAN@ENGIN.UMICH.EDU

*University of Michigan, Dept. of Electrical Engineering and Computer Science,*
*Ann Arbor, MI 48109 USA*

## Abstract

Market price systems constitute a well-understood class of mechanisms that under certain conditions provide effective decentralization of decision making with minimal communication overhead. In a *market-oriented programming* approach to distributed problem solving, we derive the activities and resource allocations for a set of computational agents by computing the competitive equilibrium of an artificial economy. WALRAS provides basic constructs for defining computational market structures, and protocols for deriving their corresponding price equilibria. In a particular realization of this approach for a form of multicommodity flow problem, we see that careful construction of the decision process according to economic principles can lead to efficient distributed resource allocation, and that the behavior of the system can be meaningfully analyzed in economic terms.

## 1. Distributed Planning and Economics

In a *distributed* or multiagent planning system, the plan for the system as a whole is a composite of plans produced by its constituent agents. These plans may interact significantly in both the resources required by each of the agents' activities (preconditions) and the products resulting from these activities (postconditions). Despite these interactions, it is often advantageous or necessary to distribute the planning process because agents are separated geographically, have different information, possess distinct capabilities or authority, or have been designed and implemented separately. In any case, because each agent has limited competence and awareness of the decisions produced by others, some sort of coordination is required to maximize the performance of the overall system. However, allocating resources via central control or extensive communication is deemed infeasible, as it violates whatever constraints dictated distribution of the planning task in the first place.

The task facing the designer of a distributed planning system is to define a computationally efficient coordination mechanism and its realization for a collection of agents. The agent configuration may be given, or may itself be a design parameter. By the term *agent*, I refer to a module that acts within the mechanism according to its own knowledge and interests. The capabilities of the agents and their organization in an overall decision-making structure determine the behavior of the system as a whole. Because it concerns the collective behavior of self-interested decision makers, the design of this decentralized structure is fundamentally an exercise in economics or incentive engineering. The problem of developing architectures for distributed planning fits within the framework of *mechanism design* (Hurwicz, 1977; Reiter, 1986), and many ideas and results from economics are directly applicable. In particular, the class of mechanisms based on price systems and competition has been deeply investigated by economists, who have characterized the conditions for its efficiency and compatibility with other features of the economy. When applicable, the competitive mechanism achieves coordination with minimal communication requirements (in a precise sense related to the dimensionality of messages transmitted among agents (Reiter, 1986)).

The theory of *general equilibrium* (Hildenbrand & Kirman, 1976) provides the foundation for a general approach to the construction of distributed planning systems based on price mechanisms. In this approach, we regard the constituent planning agents as consumers and producers in an artificial economy, and define their individual activities in terms of production and consumption of commodities. Interactions among agents are cast as exchanges, the terms of which are mediated by the underlying economic mechanism, or protocol. By specifying the universe of commodities, the configuration of agents, and the interaction protocol, we can achieve a variety of interesting and often effective decentralized behaviors. Furthermore, we can apply economic theory to the analysis of alternative architectures, and thus exploit a wealth of existing knowledge in the design of distributed planners.

I use the phrase *market-oriented programming* to refer to the general approach of deriving solutions to distributed resource allocation problems by computing the competitive equilibrium of an artificial economy.[1] In the following, I describe this general approach and a primitive programming environment supporting the specification of computational markets and derivation of equilibrium prices. An example problem in distributed transportation planning demonstrates the feasibility of decentralizing a problem with nontrivial interactions, and the applicability of economic principles to distributed problem solving.

## 2. WALRAS: A Market-Oriented Programming Environment

To explore the use of market mechanisms for the coordination of distributed planning modules, I have developed a prototype environment for specifying and simulating computational markets. The system is called WALRAS, after the 19th-century French economist Léon Walras, who was the first to envision a system of interconnected markets in price equilibrium. WALRAS provides basic mechanisms implementing various sorts of agents, auctions, and bidding protocols. To specify a computational economy, one defines a set of goods and instantiates a collection of agents that produce or consume those goods. Depending on the context, some of the goods or agents may be fixed exogenously, for example, they could correspond to real-world goods or agents participating in the planning process. Others might be completely artificial ones invented by the designer to decentralize the problem-solving process in a particular way. Given a market configuration, WALRAS then runs these agents to determine an equilibrium allocation of goods and activities. This distribution of goods and activities constitutes the market solution to the planning problem.

---

1. The name was inspired by Shoham's use of *agent-oriented programming* to refer to a specialization of object-oriented programming where the entities are described in terms of agent concepts and interact via speech acts (Shoham, 1993). Market-oriented programming is an analogous specialization, where the entities are economic agents that interact according to market concepts of production and exchange. The phrase has also been invoked by Lavoie, Baetjer, and Tulloh (1991) to refer to real markets in software components.

## 2.1 General Equilibrium

The WALRAS framework is patterned directly after general-equilibrium theory. A brief exposition, glossing over many fine points, follows; for elaboration see any text on microeconomic theory (e.g., (Varian, 1984)).

We start with $k$ goods and $n$ agents. Agents fall in two general classes. *Consumers* can buy, sell, and consume goods, and their preferences for consuming various combinations or *bundles* of goods are specified by their *utility function*. If agent $i$ is a consumer, then its utility function, $u_i : \Re_+^k \to \Re$, ranks the various bundles of goods according to preference. Consumers may also start with an initial allocation of some goods, termed their *endowment*. Let $e_{i,j}$ denote agent $i$'s endowment of good $j$, and $x_{i,j}$ the amount of good $j$ that $i$ ultimately consumes. The objective of consumer $i$ is to choose a feasible bundle of goods, $(x_{i,1}, \ldots, x_{i,k})$ (rendered in vector notation as $\mathbf{x}_i$), so as to maximize its utility. A bundle is feasible for consumer $i$ if its total cost at the going prices does not exceed the value of $i$'s endowment at these prices. The consumer's choice can be expressed as the following constrained optimization problem:

$$\max_{\mathbf{x}_i} u_i(\mathbf{x}_i) \text{ s.t. } \mathbf{p} \cdot \mathbf{x}_i \leq \mathbf{p} \cdot \mathbf{e}_i, \tag{1}$$

where $\mathbf{p} = (p_1, \ldots, p_k)$ is the vector of prices for the $k$ goods.

Agents of the second type, *producers*, can transform some sorts of goods into some others, according to their *technology*. The technology specifies the feasible combinations of inputs and outputs for the producer. Let us consider the special case where there is one output good, indexed $j$, and the remaining goods are potential inputs. In that case, the technology for producer $i$ can be described by a *production function*,

$$y_i = -x_{i,j} = f_i(x_{i,1}, \ldots, x_{i,j-1}, x_{i,j+1}, \ldots, x_{i,k}),$$

specifying the maximum output producible from the given inputs. (When a good is an input in its own production, the production function characterizes *net* output.) In this case, the producer's objective is to choose a production plan that maximizes profits subject to its technology and the going price of its output and input goods. This involves choosing a production level, $y_i$, along with the levels of inputs that can produce $y_i$ at the minimum cost. Let $\mathbf{x}_{i,\bar{j}}$ and $\mathbf{p}_{\bar{j}}$ denote the consumption and prices, respectively, of the input goods. Then the corresponding constrained optimization problem is to maximize profits, the difference between revenues and costs:

$$\max_{y_i} \left[ p_j y_i - \left[ \min_{\mathbf{x}_{i,\bar{j}}} \mathbf{p}_{\bar{j}} \cdot \mathbf{x}_{i,\bar{j}} \text{ s.t. } y_i \leq f_i(\mathbf{x}_{i,\bar{j}}) \right] \right],$$

or equivalently,

$$\min_{\mathbf{x}_i} \mathbf{p} \cdot \mathbf{x}_i \text{ s.t. } -x_{i,j} \leq f_i(\mathbf{x}_{i,\bar{j}}). \tag{2}$$

An agent acts *competitively* when it takes prices as given, neglecting any impact of its own behavior on prices. The above formulation implicitly assumes perfect competition, in that the prices are parameters of the agents' constrained optimization problems. Perfect competition realistically reflects individual rationality when there are numerous agents, each small with respect to the entire economy. Even when this is not the case, however, we can implement competitive behavior in individual agents if we so choose. The implications of the restriction to perfect competition are discussed further below.

A pair $(\mathbf{p}, \mathbf{x})$ of a price vector and vector of demands for each agent constitutes a *competitive equilibrium* for the economy if and only if:

1. For each agent $i$, $\mathbf{x}_i$ is a solution to its constrained optimization problem—(1) or (2)—at prices $\mathbf{p}$, and

2. the net amount of each good produced and consumed equals the total endowment,

$$\sum_{i=1}^{n} x_{i,j} = \sum_{i=1}^{n} e_{i,j}, \text{ for } j = 1, \ldots, k. \tag{3}$$

In other words, the total amount consumed equals the total amount produced (counted as negative quantities in the consumption bundles of producers), plus the total amount the economy started out with (the endowments).

Under certain "classical" assumptions (essentially continuity, monotonicity, and concavity of the utility and production functions; see, e.g., (Hildenbrand & Kirman, 1976; Varian, 1984)), competitive equilibria exist, and are unique given strictness of these conditions. From the perspective of mechanism design, competitive equilibria possess several desirable properties, in particular, the two fundamental welfare theorems of general equilibrium theory: (1) all competitive equilibria are *Pareto optimal* (no agent can do better without some other doing worse), and (2) *any* feasible Pareto optimum is a competitive equilibrium for some initial allocation of the endowments. These properties seem to offer exactly what we need: a bound on the quality of the solution, plus the prospect that we can achieve the most desired behavior by carefully engineering the configuration of the computational market. Moreover, in equilibrium, the prices reflect exactly the information required for distributed agents to optimally evaluate perturbations in their behavior without resorting to communication or reconsideration of their full set of possibilities (Koopmans, 1970).

## 2.2 Computing Competitive Equilibria

Competitive equilibria are also computable, and algorithms based on fixed-point methods (Scarf, 1984) and optimization techniques (Nagurney, 1993) have been developed. Both sorts of algorithms in effect operate by collecting and solving the simultaneous equilibrium equations (1), (2), and (3)). Without an expressly distributed formulation, however, these techniques may violate the decentralization considerations underlying our distributed problem-solving context. This is quite acceptable for the purposes these algorithms were originally designed, namely to analyze existing decentralized structures, such as transportation industries or even entire economies (Shoven & Whalley, 1992). But because our purpose is to *implement* a distributed system, we must obey *computational* distributivity constraints not relevant to the usual purposes of applied general-equilibrium analysis. In general, explicitly examining the space of commodity bundle allocations in the search for equilibrium undercuts our original motive for decomposing complex activities into consumption and production of separate goods.

Another important constraint is that internal details of the agents' state (such as utility or production functions and bidding policy) should be considered private in order to maximize modularity and permit inclusion of agents not under the designers' direct control. A consequence of this is that computationally exploiting global properties arising from special features of agents would not generally be permissible for our purposes. For example, the constraint that profits be zero is a consequence of competitive behavior and constant-returns production technology. Since information about the form of the technology and bidding policy is private to producer agents, it could be considered cheating to embed the zero-profit condition into the equilibrium derivation procedure.

WALRAS's procedure is a decentralized relaxation method, akin to the mechanism of *tatonnement* originally sketched by Léon Walras to explain how prices might be derived. In the basic tatonnement method, we begin with an initial vector of prices, $\mathbf{p}_0$. The agents determine their demands at those prices (by solving their corresponding constrained optimization problems), and report the quantities demanded to the "auctioneer". Based on these reports, the auctioneer iteratively adjusts the prices up or down as there is an excess of demand or supply, respectively. For instance, an adjustment proportional to the excess could be modeled by the difference equation

$$\mathbf{p}_{t+1} = \mathbf{p}_t + \alpha(\sum_{i=1}^{n} \mathbf{x}_i - \sum_{i=1}^{n} \mathbf{e}_i).$$

If the sequence $\mathbf{p}_0, \mathbf{p}_1, \ldots$ converges, then the excess demand in each market approaches zero, and the result is a competitive equilibrium. It is well known, however, that tatonnement processes do not converge to equilibrium in general (Scarf, 1984). The class of economies in which tatonnement works are those with so-called *stable* equilibria (Hicks, 1948). A sufficient condition for stability is *gross substitutability* (Arrow & Hurwicz, 1977): that if the price for one good rises, then the net demands for the other goods do not decrease. Intuitively, gross substitutability will be violated when there are *complementarities* in preferences or technologies such that reduced consumption for one good will cause reduced consumption in others as well (Samuelson, 1974).

## 2.3 WALRAS Bidding Protocol

The method employed by WALRAS successively computes an equilibrium price in each separate market, in a manner detailed below. Like tatonnement, it involves an iterative adjustment of prices based on reactions of the agents in the market. However, it differs from traditional tatonnement procedures in that (1) agents submit supply and demand *curves* rather than single point quantities for a particular price, and (2) the auction adjusts individual prices to clear, rather than adjusting the entire price vector by some increment (usually a function of summary statistics such as excess demand).[2]

WALRAS associates an *auction* with each distinct good. Agents act in the market by submitting *bids* to auctions. In WALRAS, bids specify a correspondence between prices and quantities of the good that the agent offers to demand or supply. The bid for a particular good corresponds to one dimension of the agent's optimal demand, which is parametrized by the prices for all relevant goods. Let $\mathbf{x}_i(\mathbf{p})$ be the solution to equation (1) or (2), as appropriate, for prices $\mathbf{p}$. A WALRAS agent bids for good $j$ under the assumption that prices for the remaining goods are fixed at their current values, $\mathbf{p}_{\bar{j}}$. Formally, agent $i$'s bid for good $j$ is a function $x_{i,j} : \Re_+ \rightarrow \Re$, from prices to quantities satisfying

$$x_{i,j}(p_j) = \mathbf{x}_i(p_j, \mathbf{p}_{\bar{j}})_j,$$

where the subscript $j$ on the right-hand side selects the quantity demanded of good $j$ from the overall demand vector. The agent computes and sends this function (encoded in any of a variety of formats) to the auction for good $j$.

---

2. This general approach is called *progressive equilibration* by Dafermos and Nagurney (1989), who applied it to a particular transportation network equilibrium problem. Although this model of market dynamics does not appear to have been investigated very extensively in general-equilibrium theory, it does seem to match the kind of price adjustment process envisioned by Hicks in his pioneering study of dynamics and stability (Hicks, 1948).

Given bids from all interested agents, the auction derives a market-clearing price, at which the quantity demanded balances that supplied, within some prespecified tolerance. This clearing price is simply the zero crossing of the *aggregate demand* function, which is the sum of the demands from all agents. Such a zero crossing will exist as long as the aggregate demand is sufficiently well-behaved, in particular, if it is continuous and decreasing in price. Gross substitutability, along with the classical conditions for existence of equilibrium, is sufficient to ensure the existence of a clearing price at any stage of the bidding protocol. WALRAS calculates the zero crossing of the aggregate demand function via binary search. If aggregate demand is not well-behaved, the result of the auction may be a non-clearing price.

When the current price is clearing with respect to the current bids, we say the market for that commodity is in equilibrium. We say that an agent is in equilibrium if its set of outstanding bids corresponds to the solution of its optimization problem at the going prices. If all the agents and commodity markets are in equilibrium, the allocation of goods dictated by the auction results is a competitive equilibrium.

Figure 1 presents a schematic view of the WALRAS bidding process. There is an auction for each distinct good, and for each agent, a link to all auctions in which it has an interest. There is also a "tote board" of current prices, kept up-to-date by the various auctions. In the current implementation the tote board is a global data structure, however, since price change notifications are explicitly transmitted to interested agents, this central information could be easily dispensed with.

Each agent maintains an agenda of bid tasks, specifying the markets in which it must update its bid or compute a new one. In Figure 1, agent $A_i$ has pending tasks to submit bids to auctions $G_1$, $G_7$, and $G_4$. The bidding process is highly distributed, in that each agent need communicate directly only with the auctions for the goods of interest (those in the domain of its utility or production function, or for which it has nonzero endowments). Each of these interactions concerns only a single good; auctions never coordinate with each other. Agents need not negotiate directly with other agents, nor even know of each other's existence.

As new bids are received at auction, the previously computed clearing price becomes obsolete. Periodically, each auction computes a new clearing price (if any new or updated bids have been received) and posts it on the tote board. When a price is updated, this may invalidate some of an agent's outstanding bids, since these were computed under the assumption that prices for remaining goods were fixed at previous values. On finding out

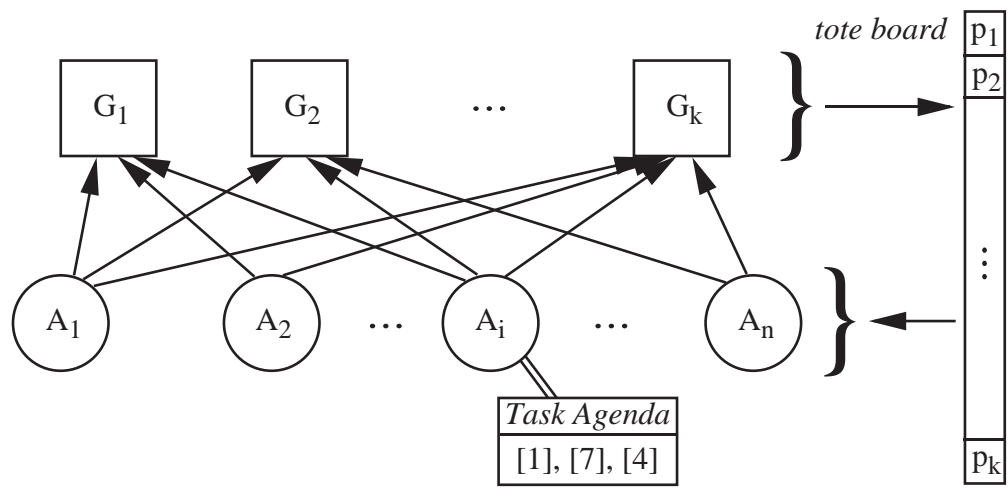

Figure 1: WALRAS's bidding process. $G_j$ denotes the auction for the $j$th good, and $A_i$ the $i$th trading agent. An item $[j]$ on the task agenda denotes a pending task to compute and submit a bid for good $j$.

about a price change, an agent augments its task agenda to include the potentially affected bids.

At all times, WALRAS maintains a vector of going prices and quantities that would be exchanged at those prices. While the agents have nonempty bid agendas or the auctions new bids, some or all goods may be in disequilibrium. When all auctions clear and all agendas are exhausted, however, the economy is in competitive equilibrium (up to some numeric tolerance). Using a recent result of Milgrom and Roberts (1991, Theorem 12), it can be shown that the condition sufficient for convergence of tatonnement—gross substitutability— is also sufficient for convergence of WALRAS's price-adjustment process. The key observation is that in progressive equilibration (synchronous or not) the price at each time is based on some set of previous supply and demand bids.

Although I have no precise results to this effect, the computational effort required for convergence to a fixed tolerance seems highly sensitive to the number of goods, and much less so to the number of agents. Eydeland and Nagurney (1989) have analyzed in detail the convergence pattern of progressive equilibration algorithms related to WALRAS for particular special cases, and found roughly linear growth in the number of agents. However, general conclusions are difficult to draw as the cost of computing the equilibrium for a particular computational economy may well depend on the interconnectedness and strength of interactions among agents and goods.

## 2.4 Market-Oriented Programming

As described above, WALRAS provides facilities for specifying market configurations and computing their competitive equilibrium. We can also view WALRAS as a programming environment for decentralized resource allocation procedures. The environment provides constructs for specifying various sorts of agents and defining their interactions via their relations to common commodities. After setting up the initial configuration, the market can be run to determine the equilibrium level of activities and distribution of resources throughout the economy.

To cast a distributed planning problem as a market, one needs to identify (1) the goods traded, (2) the agents trading, and (3) the agents' bidding behavior. These design steps are serially dependent, as the definition of what constitutes an exchangeable or producible commodity severely restricts the type of agents that it makes sense to include. And as mentioned above, sometimes we have to take as fixed some real-world agents and goods presented as part of the problem specification. Once the configuration is determined, it might be advantageous to adjust some general parameters of the bidding protocol. Below, I illustrate the design task with a WALRAS formulation of the multicommodity flow problem.

## 2.5 Implementation

WALRAS is implemented in Common Lisp and the Common Lisp Object System (CLOS). The current version provides basic infrastructure for running computational economies, including the underlying bidding protocol and a library of CLOS classes implementing a variety of agent types. The object-oriented implementation supports incremental development of market configurations. In particular, new types of agents can often be defined as slight variations on existing types, for example by modifying isolated features of the demand behavior, bidding strategies (e.g., management of task agenda), or bid format. Wang and Slagle (1993) present a detailed case for the use of object-oriented languages to represent general-equilibrium models. Their proposed system is similar to WALRAS with respect to formulation, although it is designed as an interface to conventional model-solving packages, rather than to support a decentralized computation of equilibrium directly.

Although it models a distributed system, WALRAS runs serially on a single processor. Distribution constraints on information and communication are enforced by programming and specification conventions rather than by fundamental mechanisms of the software environment. Asynchrony is simulated by randomizing the bidding sequences so that agents are called on unpredictably. Indeed, artificial synchronization can lead to an undesirable oscillation in the clearing prices, as agents collectively overcompensate for imbalances in the preceding iteration.[3]

The current experimental system runs transportation models of the sort described below, as well as some abstract exchange and production economies with parametrized utility and production functions (including the expository examples of Scarf (1984) and Shoven and Whalley (1984)). Customized tuning of the basic bidding protocol has not been necessary. In the process of getting WALRAS to run on these examples, I have added some generically useful building blocks to the class libraries, but much more is required to fill out a comprehensive taxonomy of agents, bidding strategies, and auction policies.

---

3. In some formal dynamic models (Huberman, 1988; Kephart, Hogg, & Huberman, 1989), homogeneous agents choose instantaneously optimal policies without accounting for others that are simultaneously making the same choice. Since the value of a particular choice varies inversely with the number of agents choosing it, this delayed feedback about the others' decisions leads to systematic errors, and hence oscillation. I have also observed this phenomenon empirically in a synchronized version of WALRAS. By eliminating the synchronization, agents tend to work on different markets at any one time, and hence do not suffer as much from delayed feedback about prices.

## 3. Example: Multicommodity Flow

In a simple version of the multicommodity flow problem, the task is to allocate a given set of cargo movements over a given transportation network. The transportation network is a collection of locations, with links (directed edges) identifying feasible transportation operations. Associated with each link is a specification of the cost of moving cargo along it. We suppose further that the cargo is homogeneous, and that amounts of cargo are arbitrarily divisible. A movement requirement associates an amount of cargo with an origin-destination pair. The planning problem is to determine the amount to transport on each link in order to move all the cargo at the minimum cost. This simplification ignores salient aspects of real transportation planning. For instance, this model is completely atemporal, and is hence more suitable for planning steady-state flows than for planning dynamic movements.

A distributed version of the problem would decentralize the responsibility for transporting separate cargo elements. For example, planning modules corresponding to geographically or organizationally disparate units might arrange the transportation for cargo within their respective spheres of authority. Or decision-making activity might be decomposed along hierarchical levels of abstraction, gross functional characteristics, or according to any other relevant distinction. This decentralization might result from real distribution of authority within a human organization, from inherent informational asymmetries and communication barriers, or from modularity imposed to facilitate software engineering.

Consider, for example, the abstract transportation network of Figure 2, taken from Harker (1988). There are four locations, with directed links as shown. Consider two movement requirements. The first is to transport cargo from location 1 to location 4, and the second in the reverse direction. Suppose we wish to decentralize authority so that separate agents (called shippers) decide how to allocate the cargo for each movement. The first shipper decides how to split its cargo units between the paths $1 \rightarrow 2 \rightarrow 4$ and $1 \rightarrow 2 \rightarrow 3 \rightarrow 4$, while the second figures the split between paths $4 \rightarrow 2 \rightarrow 1$ and $4 \rightarrow 2 \rightarrow 3 \rightarrow 1$. Note that the latter paths for each shipper share a common resource: the link $2 \rightarrow 3$.

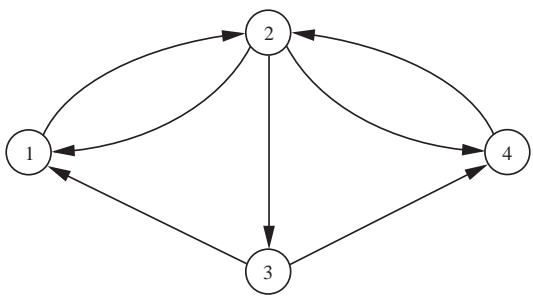

Figure 2: A simple network (from Harker (1988)).

Because of their overlapping resource demands, the shippers' decisions appear to be necessarily intertwined. In a congested network, for example, the cost for transporting a unit of cargo over a link is increasing in the overall usage of the link. A shipper planning its cargo movements as if it were the only user on a network would thus underestimate its costs and potentially misallocate transportation resources.

For the analysis of networks such as this, transportation researchers have developed equilibrium concepts describing the collective behavior of the shippers. In a *system equilibrium*, the overall transportation of cargo proceeds as if there were an omniscient central planner directing the movement of each shipment so as to minimize the total aggregate cost of meeting the requirements. In a *user equilibrium*, the overall allocation of cargo movements is such that each shipper minimizes its own total cost, sharing proportionately the cost of shared resources. The system equilibrium is thus a global optimum, while the user equilibrium corresponds to a composition of locally optimal solutions to subproblems. There are also some intermediate possibilities, corresponding to game-theoretic equilibrium concepts such as the Nash equilibrium, where each shipper behaves optimally given the transportation policies of the remaining shippers (Harker, 1986).[4]

From our perspective as designer of the distributed planner, we seek a decentralization mechanism that will reach the system equilibrium, or come as close as possible given the distributed decision-making structure. In general, however, we cannot expect to derive a system equilibrium or globally optimal solution without central control. Limits on coordination and communication may prevent the distributed resource allocation from exploiting all opportunities and inhibiting agents from acting at cross purposes. But under certain conditions decision making can indeed be decentralized effectively via market mechanisms. General-equilibrium analysis can help us to recognize and take advantage of these opportunities.

Note that for the multicommodity flow problem, there is an effective distributed solution due to Gallager (1977). One of the market structures described below effectively mimics this solution, even though Gallager's algorithm was not formulated expressly in market terms. The point here is not to crack a hitherto unsolved distributed optimization problem (though that would be nice), but rather to illustrate a general approach on a simply described yet nontrivial task.

## 4. WALRAS Transportation Market

In this section, I present a series of three transportation market structures implemented in WALRAS. The first and simplest model comprises the basic transportation goods and shipper agents, which are augmented in the succeeding models to include other agent types. Comparative analysis of the three market structures reveals the qualitatively distinct economic and computational behaviors realized by alternate WALRAS configurations.

### 4.1 Basic Shipper Model

The resource of primary interest in the multicommodity flow problem is movement of cargo. Because the value and cost of a cargo movement depends on location, we designate as a distinct good the capacity on each origin-destination pair in the network (see Figure 2). To capture the cost or input required to move cargo, we define another good denoting generic transportation resources. In a more concrete model, these might consist of vehicles, fuel, labor, or other factors contributing to transportation.

---

4. In the Nash solution, shippers correctly anticipate the effect of their own cargo movements on the average cost on each link. The resulting equilibrium converges to the user equilibrium as the number of shippers increases and the effect of any individual's behavior on prices diminishes (Haurie & Marcotte, 1985).

To decentralize the decision making, we identify each movement requirement with a distinct shipper agent. These shippers, or consumers, have an interest in moving various units of cargo between specified origins and destinations.

The interconnectedness of agents and goods defines the market configuration. Figure 3 depicts the WALRAS configuration for the basic shipper model corresponding to the example network of Figure 2. In this model there are two shippers, $S_{1,4}$ and $S_{4,1}$, where $S_{i,j}$ denotes a shipper with a requirement to move goods from origin $i$ to destination $j$. Shippers connect to goods that might serve their objectives: in this case, movement along links that belong to some simple path from the shipper's origin to its destination. In the diagram, $G_{i,j}$ denotes the good representing an amount of cargo moved over the link $i \to j$. $G_0$ denotes the special transportation resource good. Notice that the only goods of interest to both shippers are $G_0$, for which they both have endowments, and $G_{2,3}$, transportation on the link serving both origin-destination pairs.

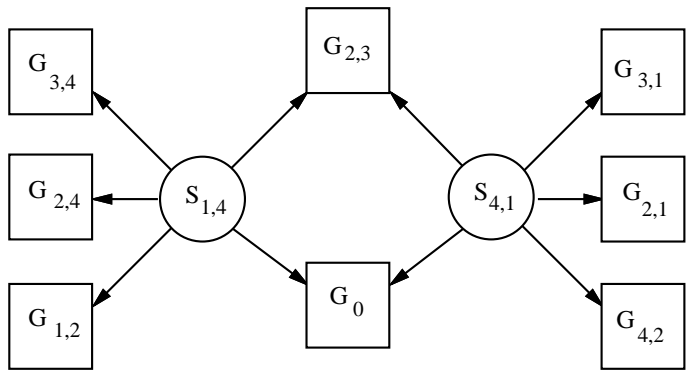

Figure 3: WALRAS basic shipper market configuration for the example transportation network.

The model we employ for transportation costs is based on a network with congestion, thus exhibiting diseconomies of scale. In other words, the marginal and average costs (in terms of transportation resources required) are both increasing in the level of service on a link. Using Harker's data, we take costs to be quadratic. The quadratic cost model is posed simply for concreteness, and does not represent any substantive claim about transportation networks. The important qualitative feature of this model (and the only one necessary for the example to work) is that it exhibits decreasing returns, a defining characteristic of congested networks. Note also that Harker's model is in terms of monetary costs, whereas we introduce an abstract input good.

Let $c_{i,j}(x)$ denote the cost in transportation resources (good $G_0$) required to transport $x$ units of cargo on the link from $i$ to $j$. The complete cost functions are:

$$c_{1,2}(x) = c_{2,1}(x) = c_{2,4}(x) = c_{4,2}(x) = x^2 + 20x,$$

$$c_{3,1}(x) = c_{2,3}(x) = c_{3,4}(x) = 2x^2 + 5x.$$

Finally, each shipper's objective is to transport 10 units of cargo from its origin to its destination.

In the basic shipper model, we assume that the shippers pay proportionately (in units of $G_0$) for the total cost on each link. This amounts to a policy of average cost pricing. We take the shipper's objective to be to ship as much as possible (up to its movement requirement) in the least costly manner. Notice that this objective is not expressible in terms of the consumer's optimization problem, equation (1), and hence this model is not technically an instance of the general-equilibrium framework.

Given a network with prices on each link, the cheapest cargo movement corresponds to the shortest path in the graph, where distances are equated with prices. Thus, for a given link, a shipper would prefer to ship its entire quota on the link if it is on the shortest path, and zero otherwise. In the case of ties, it is indifferent among the possible allocations. To bid on link $i, j$, the shipper can derive the threshold price that determines whether the link is on a shortest path by taking the difference in shortest-path distance between the networks where link $i, j$'s distance is set to zero and infinity, respectively.

In incrementally changing its bids, the shipper should also consider its outstanding bids and the current prices. The value of reserving capacity on a particular link is zero if it cannot get service on the other links on the path. Similarly, if it is already committed to shipping cargo on a parallel path, it does not gain by obtaining more capacity (even at a lower price) until it withdraws these other bids.[5] Therefore, the actual demand policy of a shipper is to spend its uncommitted income on the potential flow increase (derived from maximum-flow calculations) it could obtain by purchasing capacity on the given link. It is willing to spend up to the threshold value of the link, as described above. This determines one point on its demand curve. If it has some unsatisfied requirement and uncommitted income it also indicates a willingness to pay a lower price for a greater amount of capacity. Boundary points such as this serve to bootstrap the economy; from the initial conditions it is typically the case that no individual link contributes to overall flow between the shipper's origin and destination. Finally, the demand curve is completed by a smoothing operation on these points.

Details of the boundary points and smoothing operation are rather arbitrary, and I make no claim that this particular bidding policy is ideal or guaranteed to work for a broad class of problems. This crude approach appears sufficient for the present example and some similar ones, as long as the shippers' policies become more accurate as the prices approach equilibrium.

WALRAS successfully computes the competitive equilibrium for this example, which in the case of the basic shipper model corresponds to a user equilibrium (UE) for the transportation network. In the UE for the example network, each shipper sends 2.86 units of cargo over the shared link $2 \rightarrow 3$, and the remaining cargo over the direct link from location 2 to the destination. This allocation is inefficient, as its total cost is 1143 resource units, which is somewhat greater than the global minimum-cost solution of 1136 units. In economic terms, the cause of the inefficiency is an externality with respect to usage of the shared link. Because the shippers are effectively charged average cost—which in the case of decreasing returns is below marginal cost—the price they face does not reflect the full incremental social cost of additional usage of the resource. In effect, incremental usage of the resource by one agent is subsidized by the other. The steeper the decreasing returns, the more the agents have an incentive to overutilize the resource.[6] This is a simple example of the classic *tragedy of the commons*.

---

5. Even if a shipper could simultaneously update its bids in all markets, it would not be a good idea to do so here. A competitive shipper would send all its cargo on the least costly path, neglecting the possibility that this demand may increase the prices so that it is no longer cheapest. The outstanding bids provide some sensitivity to this effect, as they are functions of price. But they cannot respond to changes in many prices at once, and thus the policy of updating all bids simultaneously can lead to perpetual oscillation. For example, in the network considered here, the unique competitive equilibrium has each shipper splitting its cargo between two different paths. Policies allocating all cargo to one path can never lead to this result, and hence convergence to competitive equilibrium depends on the incrementality of bidding behavior.

The classical remedy to such problems is to internalize the externality by allocating ownership of the shared resource to some decision maker who has the proper incentives to use it efficiently. We can implement such a solution in WALRAS by augmenting the market structure with another type of agent.

## 4.2 Carrier Agents

We extend the basic shipper model by introducing *carriers*, agents of type producer who have the capability to transport cargo units over specified links, given varying amounts of transportation resources. In the model described here, we associate one carrier with each available link. The production function for each carrier is simply the inverse of the cost function described above. To achieve a global movement of cargo, shippers obtain transportation services from carriers in exchange for the necessary transportation resources.

Let $C_{i,j}$ denote the carrier that transports cargo from location $i$ to location $j$. Each carrier $C_{i,j}$ is connected to the auction for $G_{i,j}$, its output good, along with $G_0$—its input in the production process. Shipper agents are also connected to $G_0$, as they are endowed with transportation resources to exchange for transportation services. Figure 4 depicts the WALRAS market structure when carriers are included in the economy.

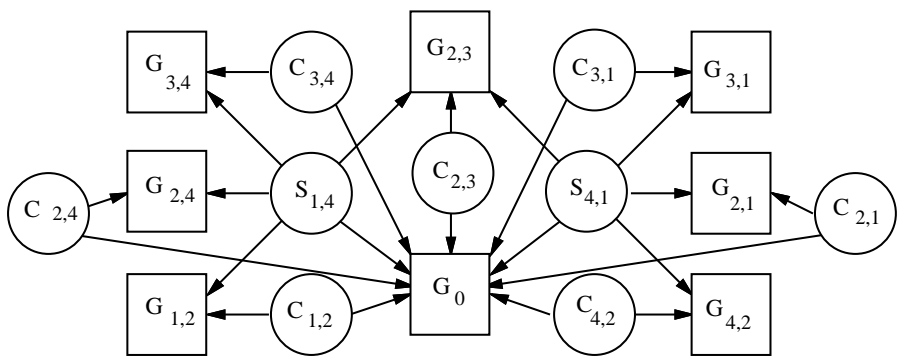

Figure 4: WALRAS market configuration for the example transportation network in an economy with shippers and carriers.

---

In the case of a decreasing returns technology, the producer's (carrier's) optimization problem has a unique solution. The optimal level of activity maximizes revenues minus costs, which occurs at the point where the output price equals marginal cost. Using this result, carriers submit supply bids specifying transportation services as a function of link prices (with resource price fixed), and demand bids specifying required resources as a function of input prices (for activity level computed with output price fixed).

For example, consider carrier $C_{1,2}$. At output price $p_{1,2}$ and input price $p_0$, the carrier's profit is

$$p_{1,2}y - p_0 c_{1,2}(y),$$

where $y$ is the level of service it chooses to supply. Given the cost function above, this expression is maximized at $y = (p_{1,2} - 20p_0)/2p_0$. Taking $p_0$ as fixed, the carrier submits a supply bid with $y$ a function of $p_{1,2}$. On the demand side, the carrier takes $p_{1,2}$ as fixed and submits a demand bid for enough good $G_0$ to produce $y$, where $y$ is treated as a function of $p_0$.

With the revised configuration and agent behaviors described, WALRAS derives the system equilibrium (SE), that is, the cargo allocation minimizing overall transportation costs. The derived cargo movements are correct to within 10% in 36 bidding cycles, and to 1% in 72, where in each cycle every agent submits an average of one bid to one auction. The total cost (in units of $G_0$), its division between shippers' expenditures and carriers' profits, and the equilibrium prices are presented in Table 1. Data for the UE solution of the basic shipper model are included for comparison. That the decentralized process produces a global optimum is perfectly consistent with competitive behavior—the carriers price their outputs at marginal cost, and the technologies are convex.

| pricing | TC | expense | profit | $p_{1,2}$ | $p_{2,1}$ | $p_{2,3}$ | $p_{2,4}$ | $p_{3,1}$ | $p_{3,4}$ | $p_{4,2}$ |
|---|---|---|---|---|---|---|---|---|---|---|
| MC (SE) | 1136 | 1514 | 378 | 40.0 | 35.7 | 22.1 | 35.7 | 13.6 | 13.6 | 40.0 |
| AC (UE) | 1143 | 1143 | 0 | 30.0 | 27.1 | 16.3 | 27.1 | 10.7 | 10.7 | 30.0 |

Table 1: Equilibria derived by WALRAS for the transportation example. TC, MC, and AC stand for total, marginal, and average cost, respectively. TC = shipper expense − carrier profit.

As a simple check on the prices of Table 1, we can verify that $p_{2,3} + p_{3,4} = p_{2,4}$ and $p_{2,3} + p_{3,1} = p_{2,1}$. Both these relationships must hold in equilibrium (assuming all links have nonzero movements), else a shipper could reduce its cost by rerouting some cargo. Indeed, for a simple (small and symmetric) example such as this, it is easy to derive the equilibrium analytically using global equations such as these. But as argued above, it would be improper to exploit these relationships in the implementation of a truly distributed decision process.

The lesson from this exercise is that we can achieve qualitatively distinct results by simple variations in the market configuration or agent policies. From our designers' perspective, we prefer the configuration that leads to the more transportation-efficient SE. Examination of Table 1 reveals that we can achieve this result by allowing the carriers to earn nonzero profits (economically speaking, these are really rents on the fixed factor represented by the congested channel) and redistributing these profits to the shippers to cover their increased expenditures. (In the model of general equilibrium with production, consumers own shares in the producers' profits. This closes the loop so that all value is ultimately realized in consumption. We can specify these shares as part of the initial configuration, just like the endowment.) In this example, we distribute the profits evenly between the two shippers.

## 4.3 Arbitrageur Agents

The preceding results demonstrate that WALRAS can indeed implement a decentralized solution to the multicommodity flow problem. But the market structure in Figure 4 is not as distributed as it might be, in that (1) all agents are connected to $G_0$, and (2) shippers need to know about all links potentially serving their origin-destination pair. The first of these concerns is easily remedied, as the choice of a single transportation resource good was completely arbitrary. For example, it would be straightforward to consider some collection of resources (e.g., fuel, labor, vehicles), and endow each shipper with only subsets of these.

The second concern can also be addressed within WALRAS. To do so, we introduce yet another sort of producer agent. These new agents, called *arbitrageurs*, act as specialized middlemen, monitoring isolated pieces of the network for inefficiencies. An arbitrageur $A_{i,j,k}$ produces transportation from $i$ to $k$ by buying capacity from $i$ to $j$ and $j$ to $k$. Its production function simply specifies that the amount of its output good, $G_{i,k}$, is equal to the minimum of its two inputs, $G_{i,j}$ and $G_{j,k}$. If $p_{i,j} + p_{j,k} < p_{i,k}$, then its production is profitable. Its bidding policy in WALRAS is to increment its level of activity at each iteration by an amount proportional to its current profitability (or decrement proportional to the loss). Such incremental behavior is necessary for all constant-returns producers in WALRAS, as the profit maximization problem has no interior solution in the linear case.[7]

To incorporate arbitrageurs into the transportation market structure, we first create new goods corresponding to the transitive closure of the transportation network. In the example network, this leads to goods for every location pair. Next, we add an arbitrageur $A_{i,j,k}$ for every triple of locations such that (1) $i \rightarrow j$ is in the original network, and (2) there exists a path from $j$ to $k$ that does not traverse location $i$. These two conditions ensure that there is an arbitrageur $A_{i,j,k}$ for every pair $i, k$ connected by a path with more than one link, and eliminate some combinations that are either redundant or clearly unprofitable.

The revised market structure for the running example is depicted in Figure 5, with new goods and agents shaded. Some goods and agents that are inactive in the market solution have been omitted from the diagram to avoid clutter.

Notice that in Figure 5 the connectivity of the shippers has been significantly decreased, as the shippers now need be aware of only the good directly serving their origin-destination pair. This dramatically simplifies their bidding problem, as they can avoid all analysis of the price network. The structure as a whole seems more distributed, as no agent is concerned with more than three goods.

---

7. Without such a restriction on its bidding behavior, the competitive constant-returns producer would choose to operate at a level of infinity or zero, depending on whether its activity were profitable or unprofitable at the going prices (at break-even, the producer is indifferent among all levels). This would lead to perpetual oscillation, a problem noticed (and solved) by Paul Samuelson in 1949 when he considered the use of market mechanisms to solve linear programming problems (Samuelson, 1966).

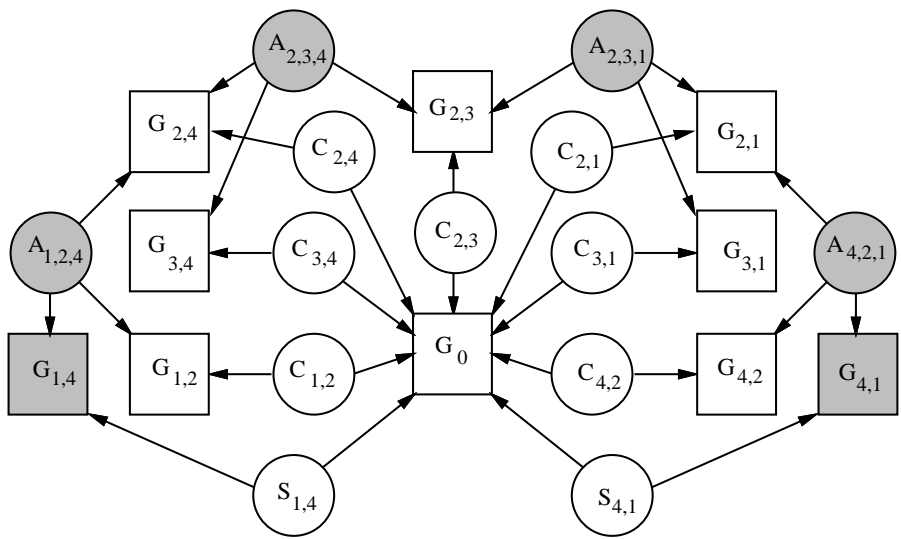

Figure 5: The revised WALRAS market configuration with arbitrageurs.

Despite the simplified shipper behavior, WALRAS still converges to the SE, or optimal solution, in this configuration. Although the resulting allocation of resources is identical, a qualitative change in market structure here corresponds to a qualitative change in the degree of decentralization.

In fact, the behavior of WALRAS on the market configuration with arbitrageurs is virtually identical to a standard distributed algorithm (Gallager, 1977) for multicommodity flow (minimum delay on communication networks). In Gallager's algorithm, distributed modules expressly differentiate the cost function to derive the marginal cost of increasing flow on a communication link. Flows are adjusted up or down so to equate the marginal costs along competing subpaths. This procedure provably converges to the optimal solution as long as the iterative adjustment parameter is sufficiently small. Similarly, convergence in WALRAS for this model requires that the arbitrageurs do not adjust their activity levels too quickly in response to profit opportunities or loss situations.

## 4.4 Summary

The preceding sections have developed three progressively elaborate market configurations for the multicommodity flow problem. Table 2 summarizes the size and shape of the configuration for a transportation network with $V$ locations and $E$ links, and $M$ movement requirements. The basic shipper model results in the user equilibrium, while both of the augmented models produce the globally optimal system equilibrium. The carrier model requires $E$ new producer agents to produce the superior result. The arbitrageur model adds $O(VE)$ more producers and potentially some new goods as well, but reduces the number of goods of interest to any individual agent from $O(E)$ to a small constant.

These market models represent three qualitatively distinct points on the spectrum of potential configurations. Hybrid models are also conceivable, for example, where a partial set of arbitrageurs are included, perhaps arranged in a hierarchy or some other regular structure. I would expect such configurations to exhibit behaviors intermediate to the specific models studied here, with respect to both equilibrium produced and degree of decentralization.

| model | goods | shippers | carriers | arbitrageurs |
|---|---|---|---|---|
| Basic shipper | $E + 1$ | $M \; [O(E)]$ | — | — |
| ...plus carriers | $E + 1$ | $M \; [O(E)]$ | $E \; [2]$ | — |
| ...plus arbitrageurs | $O(V^2)$ | $M \; [2]$ | $E \; [2]$ | $O(VE) \; [3]$ |

Table 2: Numbers of goods and agents for the three market configurations. For each type of agent, the figure in brackets indicates the number of goods on which each individual bids.

## 5. Limitations

One serious limitation of WALRAS is the assumption that agents act competitively. As mentioned above, this behavior is rational when there are many agents, each small with respect to the overall economy. However, when an individual agent is large enough to affect prices significantly (i.e., possesses market power), it forfeits utility or profits by failing to take this into account. There are two approaches toward alleviating the restriction of perfect competition in a computational economy. First, we could simply adopt models of imperfect competition, perhaps based on specific forms of imperfection (e.g., spatial monopolistic competition) or on general game-theoretic models. Second, as architects we can configure the markets to promote competitive behavior. For example, decreasing the agent's grain size and enabling free entry of agents should enhance the degree of competition. Perhaps most interestingly, by controlling the agents' knowledge of the market structure (via standard information-encapsulation techniques), we can degrade their ability to exploit whatever market power they possess. Uncertainty has been shown to increase competitiveness among risk-averse agents in some formal bidding models (McAfee & McMillan, 1987), and in a computational environment we have substantial control over this uncertainty.

The existence of competitive equilibria and efficient market allocations also depends critically on the assumption of nonincreasing returns to scale. Although congestion is a real factor in transportation networks, for example, for many modes of transport there are often other economies of scale and density that may lead to returns that are increasing overall (Harker, 1987). Note that strategic interactions, increasing returns, and other factors degrading the effectiveness of market mechanisms also inhibit decentralization in general, and so would need to be addressed directly in any approach.

Having cast WALRAS as a general environment for distributed planning, it is natural to ask how universal "market-oriented programming" is as a computational paradigm. We can characterize the computational power of this model easily enough, by correspondence to the class of convex programming problems represented by economies satisfying the classical conditions. However, the more interesting issue is how well the conceptual framework of market equilibrium corresponds to the salient features of distributed planning problems. Although it is too early to make a definitive assertion about this, it seems clear that many planning tasks are fundamentally problems in resource allocation, and that the units of distribution often correspond well with units of agency. Economics has been the most prominent (and arguably the most successful) approach to modeling resource allocation with decentralized decision making, and it is reasonable to suppose that the concepts economists find useful in the social context will prove similarly useful in our analogous computational context. Of course, just as economics is not ideal for analyzing all aspects of social interaction, we should expect that many issues in the organization of distributed planning will not be well accounted-for in this framework.

Finally, the transportation network model presented here is a highly simplified version of the actual planning problem for this domain. A more realistic treatment would cover multiple commodity types, discrete movements, temporal extent, hierarchical network structure, and other critical features of the problem. Some of these may be captured by incremental extensions to the simple model, perhaps applying elaborations developed by the transportation science community. For example, many transportation models (including Harker's more elaborate formulation (Harker, 1987)) allow for variable supply and demand of the commodities and more complex shipper-carrier relationships. Concepts of spatial price equilibrium, based on markets for commodities in each location, seem to offer the most direct approach toward extending the transportation model within WALRAS.

## 6. Related Work

### 6.1 Distributed Optimization

The techniques and models described here obviously build on much work in economics, transportation science, and operations research. The intended research contribution here is not to these fields, but rather in their application to the construction of a computational framework for decentralized decision making in general. Nevertheless, a few words are in order regarding the relation of the approach described here to extant methods for distributed optimization.

Although the most elaborate WALRAS model is essentially equivalent to existing algorithms for distributed multicommodity flow (Bertsekas & Tsitsiklis, 1989; Gallager, 1977), the market framework offers an approach toward extensions beyond the strict scope of this particular optimization problem. For example, we could reduce the number of arbitrageurs, and while this would eliminate the guarantees of optimality, we might still have a reasonable expectation for graceful degradation. Similarly, we could realize conceptual extensions to the structure of the problem, such as distributed production of goods in addition to transportation, by adding new types of agents. For any given extension, there may very well be a customized distributed optimization algorithm that would outperform the computational market, but coming up with this algorithm would likely involve a completely new analysis. Nevertheless, it must be stated that speculations regarding the methodological advantages of the market-oriented framework are indeed just speculations at this point, and the relative flexibility of applications programming in this paradigm must ultimately be demonstrated empirically.

Finally, there is a large literature on decomposition methods for mathematical programming problems, which is perhaps the most common approach to distributed optimization. Many of these techniques can themselves be interpreted in economic terms, using the close relationship between prices and Lagrange multipliers. Again, the main distinction of the approach advocated here is conceptual. Rather than taking a global optimization problem and decentralizing it, our aim is to provide a framework for formulating a task in a distributed manner in the first place.

## 6.2 Market-Based Computation

The basic idea of applying economic mechanisms to coordinate distributed problem solving is not new to the AI community. Starting with the contract net (Davis & Smith, 1983), many have found the metaphor of markets appealing, and have built systems organized around markets or market-like mechanisms (Malone, Fikes, Grant, & Howard, 1988). The original contract net actually did not include any economic notions at all in its bidding mechanism, however, recent work by Sandholm (1993) has shown how cost and price can be incorporated in the contract net protocol to make it more like a true market mechanism. Miller and Drexler (Drexler & Miller, 1988; Miller & Drexler, 1988) have examined the market-based approach in depth, presenting some underlying rationale and addressing specific issues salient in a computational environment. Waldspurger, Hogg, Huberman, Kephart, and Stornetta (1992) investigated the concepts further by actually implementing market mechanisms to allocate computational resources in a distributed operating system. Researchers in distributed computing (Kurose & Simha, 1989) have also applied specialized algorithms based on economic analyses to specific resource-allocation problems arising in distributed systems. For further remarks on this line of work, see (Wellman, 1991).

Recently, Kuwabara and Ishida (1992) have experimented with demand adjustment methods for a task very similar to the multicommodity flow problem considered here. One significant difference is that their method would consider each path in the network as a separate resource, whereas the market structures here manipulate only links or location pairs. Although they do not cast their system in a competitive-equilibrium framework, the results are congruent with those obtained by WALRAS.

WALRAS is distinct from these prior efforts in two primary respects. First, it is constructed expressly in terms of concepts from general equilibrium theory, to promote mathematical analysis of the system and facilitate the application of economic principles to architectural design. Second, WALRAS is designed to serve as a general programming environment for implementing computational economies. Although not developed specifically to allocate computational resources, there is no reason these could not be included in market structures configured for particular application domains. Indeed, the idea of grounding measures of the value of computation in real-world values (e.g., cargo movements) follows naturally from the general-equilibrium view of interconnected markets, and is one of the more exciting prospects for future applications of WALRAS to distributed problem-solving.

Organizational theorists have studied markets as mechanisms for coordinating activities and allocating resources within firms. For example, Malone (1987) models information requirements, flexibility and other performance characteristics of a variety of market and non-market structures. In his terminology, WALRAS implements a *centralized market*, where the allocation of each good is mediated by an auction. Using such models, we can determine whether this gross form of organization is advantageous, given information about the cost of communication, the flexibility of individual modules, and other related features. In this paper, we examine in greater detail the coordination process in computational markets, elaborating on the criteria for designing decentralized allocation mechanisms. We take the distributivity constraint as exogenously imposed; when the constraint is relaxable, both organizational and economic analysis illuminate the tradeoffs underlying the mechanism design problem.

Finally, market-oriented programming shares with Shoham's *agent-oriented programming* (Shoham, 1993) the view that distributed problem-solving modules are best designed and understood as rational agents. The two approaches support different agent operations (transactions versus speech acts), adopt different rationality criteria, and emphasize different agent descriptors, but are ultimately aimed at achieving the same goal of specifying complex behavior in terms of agent concepts (e.g., belief, desire, capability) and social organizations. Combining individual rationality with laws of social interaction provides perhaps the most natural approach to generalizing Newell's "knowledge level analysis" idea (Newell, 1982) to distributed computation.

## 7. Conclusion

In summary, WALRAS represents a general approach to the construction and analysis of distributed planning systems, based on general equilibrium theory and competitive mechanisms. The approach works by deriving the competitive equilibrium corresponding to a particular configuration of agents and commodities, specified using WALRAS's basic constructs for defining computational market structures. In a particular realization of this approach for a simplified form of distributed transportation planning, we see that qualitative differences in economic structure (e.g., cost-sharing among shippers versus ownership of shared resources by profit-maximizing carriers) correspond to qualitatively distinct behaviors (user versus system equilibrium). This exercise demonstrates that careful design of the distributed decision structure according to economic principles can sometimes lead to effective decentralization, and that the behaviors of alternative systems can be meaningfully analyzed in economic terms.

The contribution of the work reported here lies in the idea of market-oriented programming, an algorithm for distributed computation of competitive equilibria of computational economies, and an initial illustration of the approach on a simple problem in distributed resource allocation. A great deal of additional work will be required to understand the precise capabilities and limitations of the approach, and to establish a broader methodology for configuration of computational economies.

## Acknowledgements

This paper is a revised and extended version of (Wellman, 1992). I have benefited from discussions of computational economies with many colleagues, and would like to thank in particular Jon Doyle, Ed Durfee, Eli Gafni, Daphne Koller, Tracy Mullen, Anna Nagurney,

Scott Shenker, Yoav Shoham, Hal Varian, Carl Waldspurger, Martin Weitzman, and the anonymous reviewers for helpful comments and suggestions.

## References

Arrow, K. J., & Hurwicz, L. (Eds.). (1977). *Studies in Resource Allocation Processes*. Cambridge University Press, Cambridge.

Bertsekas, D. P., & Tsitsiklis, J. N. (1989). *Parallel and Distributed Computation*. Prentice-Hall, Englewood Cliffs, NJ.

Dafermos, S., & Nagurney, A. (1989). Supply and demand equilibration algorithms for a class of market equilibrium problems. *Transportation Science*, *23*, 118–124.

Davis, R., & Smith, R. G. (1983). Negotiation as a metaphor for distributed problem solving. *Artificial Intelligence*, *20*, 63–109.

Drexler, K. E., & Miller, M. S. (1988). Incentive engineering for computational resource management. In Huberman (1988), pp. 231–266.

Eydeland, A., & Nagurney, A. (1989). Progressive equilibration algorithms: The case of linear transaction costs. *Computer Science in Economics and Management*, *2*, 197–219.

Gallager, R. G. (1977). A minimum delay routing algorithm using distributed computation. *IEEE Transactions on Communications*, *25*, 73–85.

Harker, P. T. (1986). Alternative models of spatial competition. *Operations Research*, *34*, 410–425.

Harker, P. T. (1987). *Predicting Intercity Freight Flows*. VNU Science Press, Utrecht, The Netherlands.

Harker, P. T. (1988). Multiple equilibrium behaviors on networks. *Transportation Science*, *22*, 39–46.

Haurie, A., & Marcotte, P. (1985). On the relationship between Nash-Cournot and Wardrop equilibria. *Networks*, *15*, 295–308.

Hicks, J. R. (1948). *Value and Capital* (second edition). Oxford University Press, London.

Hildenbrand, W., & Kirman, A. P. (1976). *Introduction to Equilibrium Analysis: Variations on Themes by Edgeworth and Walras*. North-Holland Publishing Company, Amsterdam.

Huberman, B. A. (Ed.). (1988). *The Ecology of Computation*. North-Holland.

Hurwicz, L. (1977). The design of resource allocation mechanisms. In Arrow and Hurwicz (1977), pp. 3–37. Reprinted from *American Economic Review Papers and Proceedings*, 1973.

Kephart, J. O., Hogg, T., & Huberman, B. A. (1989). Dynamics of computational ecosystems. *Physical Review A*, *40*, 404–421.

Koopmans, T. C. (1970). Uses of prices. In *Scientific Papers of Tjalling C. Koopmans*, pp. 243–257. Springer-Verlag. Originally published in the Proceedings of the Conference on Operations Research in Production and Inventory Control, 1954.

Kurose, J. F., & Simha, R. (1989). A microeconomic approach to optimal resource allocation in distributed computer systems. *IEEE Transactions on Computers*, *38*, 705–717.

Kuwabara, K., & Ishida, T. (1992). Symbiotic approach to distributed resource allocation: Toward coordinated balancing. In *Pre-Proceedings of the 4th European Workshop on Modeling Autonomous Agents in a Multi-Agent World*.

Lavoie, D., Baetjer, H., & Tulloh, W. (1991). Coping with complexity: OOPS and the economists' critique of central planning. *Hotline on Object-Oriented Technology*, *3*(1), 6–8.

Malone, T. W., Fikes, R. E., Grant, K. R., & Howard, M. T. (1988). Enterprise: A market-like task scheduler for distributed computing environments. In Huberman (1988), pp. 177–205.

Malone, T. W. (1987). Modeling coordination in organizations and markets. *Management Science*, *33*, 1317–1332.

McAfee, R. P., & McMillan, J. (1987). Auctions and bidding. *Journal of Economic Literature*, *25*, 699–738.

Milgrom, P., & Roberts, J. (1991). Adaptive and sophisticated learning in normal form games. *Games and Economic Behavior*, *3*, 82–100.

Miller, M. S., & Drexler, K. E. (1988). Markets and computation: Agoric open systems. In Huberman (1988), pp. 133–176.

Nagurney, A. (1993). *Network Economics: A Variational Inequality Approach*. Kluwer Academic Publishers.

Newell, A. (1982). The knowledge level. *Artificial Intelligence*, *18*, 87–127.

Reiter, S. (1986). Information incentive and performance in the (new)$^2$ welfare economics. In Reiter, S. (Ed.), *Studies in Mathematical Economics*. MAA Studies in Mathematics.

Samuelson, P. A. (1966). Market mechanisms and maximization. In Stiglitz, J. E. (Ed.), *The Collected Scientific Papers of Paul A. Samuelson*, Vol. 1, pp. 415–492. MIT Press, Cambridge, MA. Originally appeared in RAND research memoranda, 1949.

Samuelson, P. A. (1974). Complementarity: An essay on the 40th anniversary of the Hicks-Allen revolution in demand theory. *Journal of Economic Literature*, *12*, 1255–1289.

Sandholm, T. (1993). An implementation of the contract net protocol based on marginal cost calculations. In *Proceedings of the National Conference on Artificial Intelligence*, pp. 256–262 Washington, DC. AAAI.

Scarf, H. E. (1984). The computation of equilibrium prices. In Scarf, H. E., & Shoven, J. B. (Eds.), *Applied General Equilibrium Analysis*, pp. 1–49. Cambridge University Press, Cambridge.

Shenker, S. (1991). Congestion control in computer networks: An exercise in cost-sharing. Prepared for delivery at Annual Meeting of the American Political Science Association.

Shoham, Y. (1993). Agent-oriented programming. *Artificial Intelligence*, *60*, 51–92.

Shoven, J. B., & Whalley, J. (1984). Applied general-equilibrium models of taxation and international trade: An introduction and survey. *Journal of Economic Literature*, *22*, 1007–1051.

Shoven, J. B., & Whalley, J. (1992). *Applying General Equilibrium*. Cambridge University Press.

Varian, H. R. (1984). *Microeconomic Analysis* (second edition). W. W. Norton & Company, New York.

Waldspurger, C. A., Hogg, T., Huberman, B. A., Kephart, J. O., & Stornetta, S. (1992). Spawn: A distributed computational economy. *IEEE Transactions on Software Engineering*, *18*, 103–117.

Wang, Z., & Slagle, J. (1993). An object-oriented knowledge-based approach for formulating applied general equilibrium models. In *Third International Workshop on Artificial Intelligence in Economics and Management* Portland, OR.

Wellman, M. P. (1991). Review of Huberman (1988). *Artificial Intelligence*, *52*, 205–218.

Wellman, M. P. (1992). A general-equilibrium approach to distributed transportation planning. In *Proceedings of the National Conference on Artificial Intelligence*, pp. 282–289 San Jose, CA. AAAI.